\documentclass[sigconf]{acmart}
\usepackage{xcolor}
\usepackage{multicol}
\usepackage[compact,small]{titlesec}
\usepackage{enumitem}
\usepackage[most]{tcolorbox}
\usepackage{graphicx}

\AtBeginDocument{%
  }

\copyrightyear{2024}
\acmYear{2024}
\setcopyright{rightsretained}
\acmConference[ICCAD '24]{IEEE/ACM International Conference on Computer-Aided Design}{October 27--31, 2024}{New York, NY, USA}
\acmBooktitle{IEEE/ACM International Conference on Computer-Aided Design (ICCAD '24), October 27--31, 2024, New York, NY, USA}
\acmDOI{10.1145/3676536.3698390}
\acmISBN{979-8-4007-1077-3/24/10}

\begin{document}

\title{Generative AI Agents in Autonomous Machines:\\A Safety Perspective}

\author{Jason Jabbour}
\affiliation{%
  \institution{Harvard University}
  \city{Boston}
  \state{Massachusetts}
  \country{United States}}
\email{jasonjabbour@g.harvard.edu}

\author{Vijay Janapa Reddi}
\affiliation{%
  \institution{Harvard University}
  \city{Boston}
  \state{Massachusetts}
  \country{United States}}
\email{vj@eecs.harvard.edu}

\begin{abstract}

The integration of Generative Artificial Intelligence (AI) into autonomous machines represents a major paradigm shift in how these systems operate and unlocks new solutions to problems once deemed intractable. Although generative AI agents provide unparalleled capabilities, they also have unique safety concerns. These challenges require robust safeguards, especially for autonomous machines that operate in high-stakes environments. This work investigates the evolving safety requirements when generative models are integrated as agents into physical autonomous machines, comparing these to safety considerations in less critical AI applications. We explore the challenges and opportunities to ensure the safe deployment of generative AI-driven autonomous machines. Furthermore, we provide a forward-looking perspective on the future of AI-driven autonomous systems and emphasize the importance of evaluating and communicating safety risks. As an important step towards addressing these concerns, we recommend the development and implementation of comprehensive safety scorecards for the use of generative AI technologies in autonomous machines.

\end{abstract}

\begin{teaserfigure}
    \includegraphics[width=\textwidth]{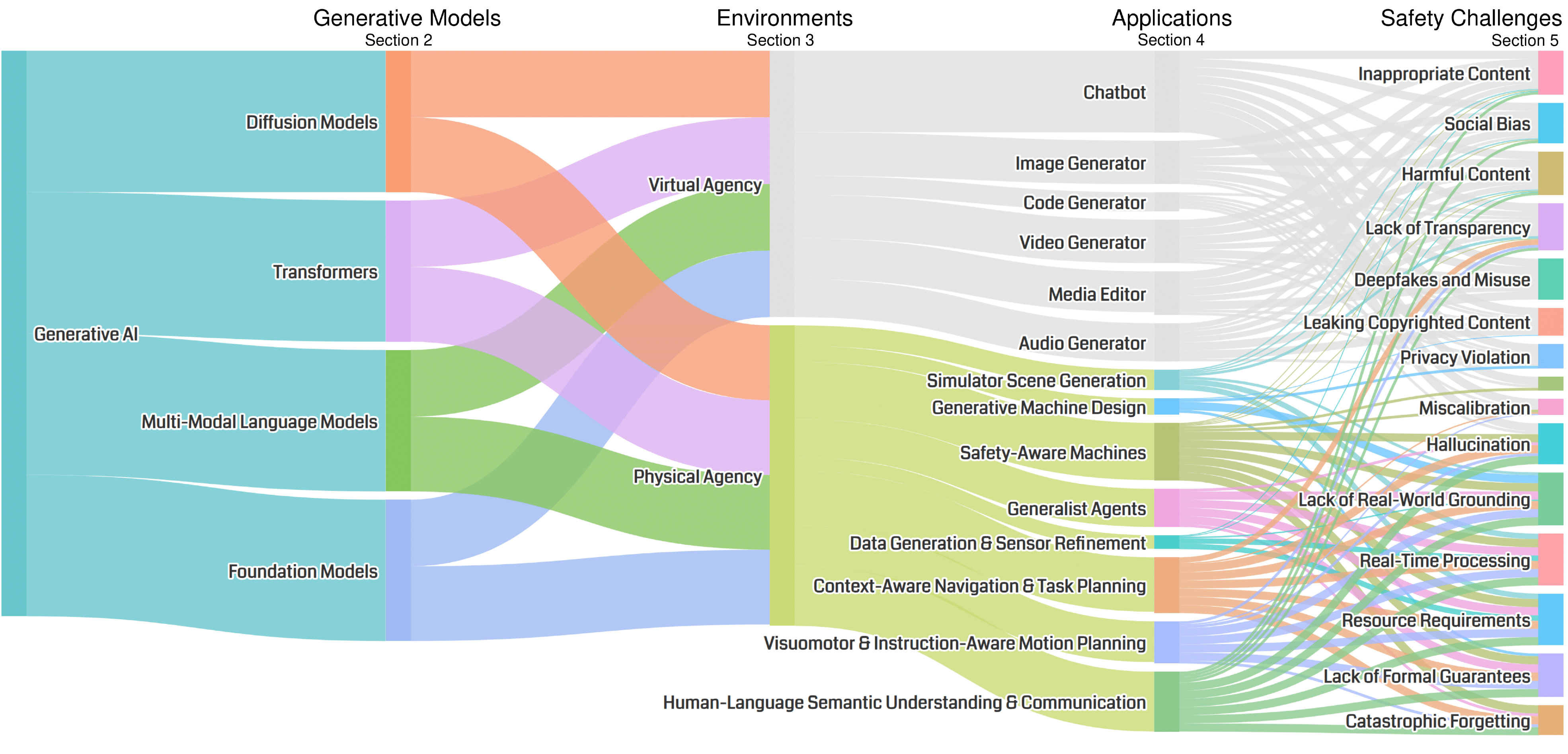}  
    \caption{Generative AI is increasingly used in autonomous machines. The transition of generative models from virtual environments (e.g. chatbots) to systems with physical agency amplifies existing safety concerns such as hallucination and harmful content generation. In addition, this shift also introduces new challenges like catastrophic forgetting, real-time processing, resource requirements, lack of formal guarantees, and lack of real-world grounding, making safety-critical considerations more pronounced in systems with physical agency.}
  \label{fig:teaser}
\end{teaserfigure}



\maketitle


\section{Introduction}

Generative Artificial Intelligence (AI) refers to models capable of generating new data by learning patterns and distributions from existing datasets. Powered by neural network architectures such as transformers \cite{vaswani2017attention} and diffusion \cite{ho2020denoising, peebles2023scalable, song2020denoising}, these models have shown remarkable proficiency in generating realistic, high-quality data. Examples include natural language processing (NLP) systems such as text generators and image synthesis models, both of which are capable of producing outputs that closely mimic human-like interactions \cite{bubeck2023sparks} and visuals \cite{ruiz2023dreambooth, saharia2022photorealistic}. Models, such as GPT-4 \cite{achiam2023gpt} and DALL-E \cite{ramesh2022hierarchical, ramesh2021zero}, expand the potential of generative AI with capabilities such as contextual understanding, cross-domain generalization, and the ability to handle various multimodal inputs. 

\begin{figure}[t]
    \centering
    \includegraphics[width=\linewidth]{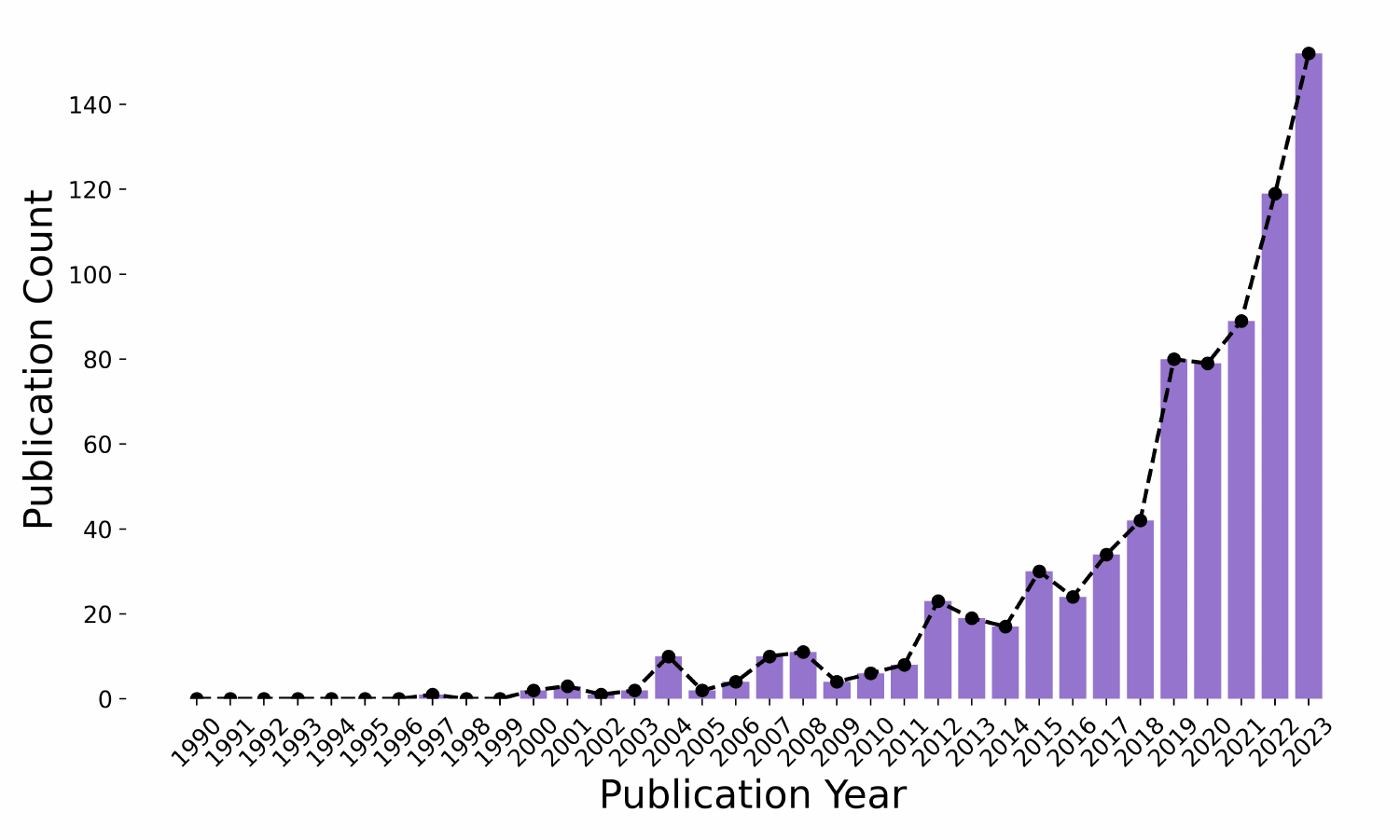}  
    \caption{Publication trend from IROS, ICRA, and RSS proceedings, using the keywords LLMs, generative models, and embodied AI. The data shows a sharp rise in papers on generative methods in robotics, especially since 2018, reflecting growing research interest.}
    \label{fig:trend}
\end{figure}

Generative AI models have become valuable in autonomous systems (e.g. self-driving cars, robotic manipulation), where their context-aware reasoning and semantic understanding drive breakthroughs in navigation, perception, and task planning \cite{belkhale2024rt, chang2023goat, chi2023diffusion, yu2023language}. The rise in publications reflects the growing integration of generative models into autonomous systems, as shown in Figure~\ref{fig:trend}. 

Generative AI has seen widespread adoption in virtual settings, which comes with safety challenges such as generating false information, often called ``hallucinations" \cite{jones2023teaching, liu2023evaluating}, or producing inappropriate visuals \cite{quaye2024adversarial}. Although these issues are significant, they generally do not pose physical safety risks since they occur within virtual environments. When generative AI is applied to systems with physical agency, such as self-driving cars, the safety stakes are significantly higher. Such systems operate in the physical world and directly affect human lives, property, and critical infrastructure. This not only amplifies the existing safety challenges faced by virtual agents but also introduces new ones. As illustrated in Figure~\ref{fig:teaser}, the transition of generative models to systems with physical agency introduces additional challenges such as catastrophic forgetting, real-time processing constraints, increased resource requirements, lack of formal guarantees, and the need for better real-world grounding. Moreover, existing problems like hallucinations and harmful content generation become more critical in physically-embodied systems, making safety-critical considerations more pronounced.

In this paper, we first provide an introduction to generative AI, outlining its fundamental principles and the types of models that have driven its success. We then survey and categorize the most recent applications of generative models in autonomous systems. Following this taxonomy, we explore the unique safety challenges posed by integrating generative AI into autonomous machines and present the latest opportunities to address these challenges. We recommend implementing a safety scorecard to enable an easy high-level assessment of generative AI models in autonomous machines.

\section{Generative AI Models Background}

We provide background on the development of generative AI models that have enabled deeper contextual understanding and reasoning capabilities. This lays the groundwork for their use cases. 

\subsection{Early Image Generating Models}

The rise of generative models began with autoencoders \cite{baldi2012autoencoders, ranzato2007unsupervised, masci2011stacked, bank2023autoencoders}, which compress data into a latent space and reconstruct inputs by minimizing reconstruction error. While useful for dimensionality reduction and denoising, they struggled with generating realistic data. Variational Autoencoders (VAEs) \cite{kingma2013auto} and their conditional variants \cite{sohn2015learning} introduced probabilistic latent spaces, enabling data generation through sampling. Despite improving realism, VAEs often produced images that lacked sharpness and detail.

A major breakthrough came with Generative Adversarial Networks (GANs) \cite{goodfellow2014generative, karras2019style} and their conditional variant \cite{mirza2014conditional}, which pit a generator and discriminator against each other in a zero-sum game to learn how to generate realistic data. GANs have excelled in tasks like image synthesis \cite{karras2019style, isola2017image, zhu2017unpaired, deng20233d}, super-resolution \cite{ledig2017photo}, and text-to-image generation \cite{reed2016generative}, often using U-Nets \cite{ronneberger2015u} for detail preservation. Despite successes, GANs suffer from unstable training and inconsistent quality \cite{arjovsky2017towards, brock2018large}. These challenges paved the way for the development of diffusion models, which offered a more robust approach to image and data generation.



\subsection{Early Sequence Generating Models}

In parallel to advancements in generative models for image data, sequence models underwent their own evolution. Recurrent Neural Networks (RNNs) \cite{rumelhart1986learning, robinson1987utility, schmidt2019recurrent} were among the first models to handle sequential data by utilizing hidden states, but they struggled with long-term dependencies \cite{bengio1993problem} and gradient issues \cite{pascanu2013difficulty}. This led to Long-Short-Term Memory (LSTM) networks \cite{hochreiter1997long, staudemeyer2019understanding}, which introduced gating mechanisms to better handle information flow across long sequences. LSTMs significantly improved handling long-term dependencies, excelling in tasks such as language translation \cite{sutskever2014sequence, wu2016google}, time-series forecasting \cite{sunny2020deep, siami2018comparison}, and music generation \cite{wu2019hierarchical}. However, they struggled with scaling to very long sequences and large datasets due to efficiency limitations \cite{kuchaiev2017factorization, shazeer2017outrageously}. This led to the development of transformers, which have since become the backbone of modern NLP and other sequence-based tasks.

\subsection{Diffusion Models}

Unlike GANs, which rely on adversarial training, diffusion models iteratively refine noisy data to produce high-fidelity outputs \cite{ho2020denoising}. Diffusion models operate in two stages: the forward (noising) process and the backward (denoising) process, both modeled as Markov chains. In the forward process, Gaussian noise is added step-by-step until the data become indistinguishable from random noise. During training, the model learns to predict the noise present and the uncertainty (variance) in this prediction at each step. Once trained, diffusion models can take random noise—either unconditioned or guided by input conditions—and progressively refine it through the denoising process to generate highly realistic outputs. This iterative refinement enables the model to transform noise into structured data, such as high-resolution images \cite{rombach2022high, hoogeboom2023simple, dhariwal2021diffusion, podell2023sdxl}, 3D models \cite{tang2023make, hu2024efficientdreamer, jun2023shap, long2024wonder3d}, or videos \cite{ho2022imagen, ho2022video, blattmann2023align}. The result is superior in stability and quality compared to previous models.

\subsection{Transformer Models}

The transformer architecture \cite{vaswani2017attention} solves the limitations of LSTMs, such as difficulties with long-term dependencies and lack of parallelization. Transformers process entire sequences in parallel for greater efficiency, adding positional embeddings to maintain the order of tokens without the need to process them sequentially. The self-attention mechanism in transformers captures relationships between tokens by calculating how much focus each token should have on others, updating their representations based on these interactions. This process helps the model understand the significance of each token, and the decoder uses this information to predict the most likely next token. Transformers have redefined sequence modeling and now serve as the backbone of cutting-edge systems in text generation \cite{guo2021longt5, liu2018generating}, language translation \cite{xue2020mt5, liu2020multilingual}, and code generation \cite{svyatkovskiy2020intellicode, li2022competition, le2022coderl, berabi2021tfix}, thanks to their ability to capture long-range dependencies and process sequences in parallel.

\subsection{Multi-Modal Language Models}

As transformer models proved highly effective in handling sequential data, their capabilities soon extended beyond text to multimodal applications, integrating different data types such as images, text, and speech. A key milestone in the shift towards multimodal models was the introduction of the Vision Transformer (ViT) \cite{dosovitskiy2020image}, which applied transformers to images by splitting them into patches treated as tokens, much like words in language models. Building on this, CLIP (Contrastive Language–Image Pretraining) \cite{radford2021learning} aligned vision and language by training on image-text pairs, associating visual data with textual descriptions. This alignment became foundational for Vision-Language Models (VLMs) \cite{caffagni2024r}, which integrate image and text inputs to generate text outputs, driving applications such as image captioning and visual question answering \cite{liu2024improved, cai2023making, alayrac2022flamingo, wang2022git}. 
Multimodal capabilities have since expanded beyond vision and language, incorporating other modalities such as audio \cite{deshmukh2023pengi}. 


\subsection{Foundation Models}

The evolution of generative models has led to the rise of foundation models, large-scale models trained on vast datasets, often across diverse modalities \cite{bai2023qwen}. These models are notable for their immense size: GPT-3, for example, has 175 billion parameters and was trained on 300 billion tokens, equivalent to 45TB of compressed plaintext and requiring 3,640 petaflop-days of compute during pretraining \cite{brown2020language}. This massive scale enables foundation models to capture intricate patterns and relationships that smaller models cannot. Fine-tuned using Reinforcement Learning from Human Feedback (RLHF) \cite{ouyang2022training}, they generalize between tasks and better align with human expectations. As a result, foundation models exhibit not only impressive generative capabilities, but also advanced contextual understanding and reasoning \cite{bubeck2023sparks, wei2022chain}, allowing them to grasp nuanced prompts and produce outputs that reflect deep comprehension of visual and textual inputs \cite{achiam2023gpt}. This ability to reason and adapt has spurred a new wave of advancements in fields such as robotics, unlocking capabilities previously unattainable.

\section{Environments}

As shown in Figure~\ref{fig:teaser} (under ``Environment''), the aforementioned generative AI models can serve in two different agentic forms: virtual and physical. In \textit{virtual agency}, the generative AI agents act in digital environments, where the action space is limited to tasks such as generating text, images, or music. These agents have control over virtual elements, but their actions do not directly influence the physical world. Although virtual agents can pose challenges like sycophancy or leaking copyright content, the impact remains confined to the digital realm. Although significant, the risks posed by virtual agency are generally confined to non-physical harms.

In contrast, \textit{physical agency} has a larger and more complex action space. Generative models in this domain have physical influence, such as moving a robotic arm, navigating a vehicle, or training an agent in simulation that will later be deployed in real-world environments. The consequences are more severe: errors can lead to physical damage, injury, or critical failures. 

Moving from virtual to physical environments greatly increases the need for safety due to direct real-world interactions. This paper focuses on physical agency, where generative AI requires higher and special standards of safety than in virtual settings.

\section{Generative AI in Autonomous Machines}

We explore the latest applications of generative AI in robotics, highlighting how models such as diffusion, transformers, and foundation models are advancing autonomous systems. Each category begins with a discussion of the pressing challenges in the area, followed by a description of generative methods that offer effective solutions.


\subsection{Simulator Scene Generation}


A key pressing challenge in the development of autonomous machines is the creation of scalable, realistic, and diverse simulation environments. Developing for the real world requires significant capital and labor and poses safety risks, making scaling impractical. So robotics turns to simulation, but one of the major challenges is the labor intensive process of creating realistic and diverse simulation environments. Simulations are difficult to control, as designing visual assets, configuring physics parameters, and defining task-specific details are time-consuming tasks that limit scalability. The manual process of identifying and modeling real-world scenes can produce high-quality results, but it often lacks the complexity and variability required for robust policy transfer, frequently leading to a sim-to-real gap. Scalable simulation content must meet three key criteria: it must be realistic enough to ensure that machine learning models trained within the simulation can transfer effectively to the real world; it must offer diversity in scenes, assets, and tasks to enable generalizable learning; and it must be easily controllable, allowing for the targeted generation of specific scenarios.
 
Generative models offer a scalable solution to many of the challenges in creating realistic and diverse simulation environments. By automating the generation of 3D assets, textures, and physics configurations from high-level inputs, these models significantly reduce manual effort, enabling the creation of diverse and scalable environments \cite{katara2024gen2sim, chen2024urdformer, nasiriany2024robocasa}. Language models have enhanced controllability by transforming high-level descriptions into specific simulation scenarios, making it possible to generate complex simulations from simple text inputs \cite{tan2023language, zhong2023language, ling2024socialgail}. Generative models also streamline sim-to-real transfer by automating the design of reward functions and simulation parameters, reducing the need for manual adjustments and improving adaptability \cite{ma2024dreureka, yu2024natural}.

\subsection{Data Generation and Sensor Refinement}

Another challenge is the acquisition of large and diverse datasets necessary for generalizable robot learning. Robot learning has great potential for generalizing across a wide range of tasks, environments, and objects, but achieving this requires large and diverse datasets, which are costly and difficult to collect in real-world settings. Unlike the abundance of text data readily available on the web, robotics depends on specialized data sources like tactile sensing for identifying and grasping objects. Collecting this data is often limited by expensive techniques like on-robot teleoperation. Furthermore, the rigidity of most autonomous machine setups makes it difficult to collect diverse data across various scenarios, leaving robotics datasets constrained to single setups with only a few hours of data. Even with access to sensors like LiDAR or RGB cameras, hardware limitations often lead to low resolution, sparse data. 

Generative models help scale robotic datasets using intelligent labeling and next-generation data augmentation techniques. Text-guided diffusion models can generate realistic images by enriching existing datasets with new objects, scenes, and annotations, while modifying textures, shapes, and backgrounds in ways that are physically consistent with real-world tasks. Pre-trained VLMs expand these datasets further by adding detailed semantic concepts, transforming basic descriptions like ``pick apple'' into more nuanced variations such as ``the red-colored fruit'', enriching the data for downstream tasks \cite{yu2023scaling, xiao2022robotic, krupnik2023fine, chen2023genaug}. Generative models also augment sensor-rich datasets, such as tactile sensory data for grasping, and expand grasp datasets with a wider variety of objects \cite{zhong2023touching, vuong2023grasp}. For more common sensory inputs, such as cameras or LiDAR, diffusion models enhance low-resolution data by upscaling it and creating higher-quality representations, improving perception in autonomous systems \cite{nakashima2024lidar, zhou20243d, richard2024omnilrs, helgesen2024fast}.


\subsection{Context-Aware Navigation \& Task Planning}

General-purpose robots has long been a goal of the robotics community, yet current systems remain brittle, often failing when confronted with unfamiliar environments. Unlike humans, who navigate the physical world using prior knowledge, robots lack the common sense understanding needed to handle everyday tasks and adapt to new settings. Humans excel at building cognitive maps, enabling them to locate landmarks, infer spatial layouts, and use semantic knowledge to navigate unfamiliar spaces. For robots to perform human-like exploration and task planning, they need to grasp how environments are organized semantically, such as recognizing that books are typically found near bookshelves in a living room. Tasks like ``cleaning the kitchen'' require not just an understanding of objects, such as knowing that a sponge is used for wiping surfaces, but also the logical sequences of actions, such as placing soap on the sponge before wiping the surface, and the ability to adapt to user preferences.

VLMs and large language models (LLMs) allow robots to integrate common sense reasoning and semantic understanding, helping them associate language commands with physical spaces and objects, enabling them to navigate and perform tasks in new environments \cite{zhang2024navid, gramopadhye2023generating, singh2023progprompt, chen2023open, yokoyama2024vlfm, huang2022inner, sun2024prompt, jain2023transformers}. These models also facilitate the creation of semantic maps that help robots build a structured understanding of their surroundings for efficient exploration, object localization, and memory of past experiences \cite{werby2024hierarchical, chang2023goat, rana2023sayplan, liu2024ok, ren2024explore, shah2023navigation, huang2023visual}. Robots can now combine multimodal sensors, to enhance physical reasoning for tasks that involve material properties, such as identifying soft objects by touch \cite{yu2024octopi}. Generative AI  allows robots to navigate based on learned semantic knowledge, inferring object relations, and spatial layouts (e.g., recognizing that certain objects are located together) \cite{biggie2023tell, sharma2023semantic, yu2023l3mvn}. 


\subsection{Visuomotor \& Instruction-Aware Motion Planning}

Humans can easily use a wide range of tools, from hammers to joysticks, with minimal instruction by understanding both the physical and cognitive affordances of objects.  Robots, however, struggle with this adaptability, especially when it comes to handling diverse, articulated objects like home appliances or furniture. Manipulating these objects requires understanding both their physical structure and functional uses, which is complicated when an object’s appearance doesn't align with its function. For example, turning on a stove by rotating a knob requires recognizing it as a control mechanism, not as an object to push or pull. This mismatch between structure and function challenges robots to bridge the gap between high-level understanding and low-level motion planning. Multimodal data, such as visual and linguistic inputs, are key to advancing their ability to adapt to human-like, lifelong learning when handling manipulation tasks, yet mapping these high-dimensional observations to low-level actions has been a long-standing challenge in robotics.

Generative models, particularly diffusion models and transformers, have demonstrated remarkable capabilities in generating complex motion trajectories, helping robots handle dynamic tasks like reorienting objects, planning precise paths, and executing 7-DoF manipulator trajectories in 3D \cite{janner2022planning, prasad2024consistency, ze20243d, xian2023chaineddiffuser, yoneda2023noise, mishra2023generative, liu2024composable, wang2023cold, mendez2023embodied, morgan2024cppflow, mishra2024reorientdiff, carvalho2023motion}. Beyond motion planning, these models have bridged the gap between high-level language inputs and low-level motor control, providing robotics with instruction-aware manipulation \cite{chen2023playfusion, yu2023language, wen2023any, geng2024sage, liu2022structdiffusion, chen2023polarnet}. Generative models enable flexible conditioning, allowing robots to integrate visual perception with control policies and handle highly multimodal data by modeling the full distribution via generative models. They can augment classic algorithms like Model Predictive Control (MPC) by giving them the "eyes" of VLMs, enhancing their ability to perceive and respond to complex environments \cite{vosylius2024render, zhao2024vlmpc, di2024keypoint, shridhar2023perceiver, chi2023diffusion, sridhar2024nomad}. LLMs have transformed how robots reason about affordances and tool use, allowing them to infer an object's function from its appearance and adapt to new tools in a given context \cite{ren2023leveraging, huang2023voxposer, birr2024autogpt+, liu2024moka}.

\subsection{Human-Language Semantic Understanding and Communication}

Natural language offers a rich interface for humans to interact with robots, enabling even those with minimal training to direct behaviors, express preferences, and provide feedback. However, interpreting and following language commands has been a significant challenge in robotics. Humans can communicate through direct instructions (e.g. ``Move the box to the corner'') or more ambiguous commands (e.g. ``We’re getting ready for a big event!''), and robots must learn to navigate this range of specificity.
To be effective in human environments, robots need to understand the semantics of language and learn from human feedback, which can include adjusting goals (e.g., ``Clean the bedroom instead''), adding constraints (e.g., ``Avoid stepping on the rug''), or offering hints when the robot is stuck. They must also learn from observations and demonstrations, much like infants learning through interaction. While natural language plays a central role in communication, human-robot collaboration also relies on non-verbal cues such as gestures. Gestures like pointing provide an efficient way to express intent, but robots must be able to accurately infer the meaning within the context of a task—a challenge that further complicates human-robot interaction.

Using LLMs, VLMs, and transformers, robots can follow natural language instructions or learn from minimal human demonstrations, significantly improving the interface between humans and robots by allowing more natural communication \cite{goyal2024rvt, jain2024vid2robot, karamcheti2023language, bandyopadhyay2024demonstrating, rivkin2023ansel}. Beyond language-based instructions, LLMs and transformers have also expanded nonverbal communication, using gestures and sketches as input to help robots infer human intent and collaborate more naturally \cite{lin2023gesture, gu2023robotic}. Building on these capabilities, generative models allow robots to interpret simple and vague instructions while grounding commands in visual and contextual cues, improving their ability to handle diverse and complex environments \cite{belkhale2024rt, tang2023saytap, liu2023grounding, sundaresan2023kite, stone2023open}. Also, robots are able to adapt to feedback in real time, refining their actions based on corrections and even asking for clarifications when instructions are ambiguous, making them more accessible to nonexperts by facilitating more clear and interactive human-robot collaborations \cite{liang2024learning, sharma2022correcting, majumdar2023findthis, han2024interpret}.

\subsection{Generative Machine Design}

Generative design is emerging in robotics, where models optimize and accelerate the design process by efficiently exploring the vast design space. Unlike time-consuming manual methods, generative models enable rapid iteration and the exploration of multiple configurations to identify optimal solutions. This is particularly useful in designing modular robots, where a set of components such as bodies, legs, and wheels must be configured for specific tasks or terrains \cite{hu2022modular}. Generative design also addresses complex tasks in manipulator and soft robot design, generating custom geometries that are highly adapted to their tasks \cite{xu2024dynamics, chan2024creation}. Recent advances have developed physics-informed diffusion models, enabling the generation of designs that not only meet aesthetic or structural criteria, but also optimize performance based on physical simulations, such as minimizing drag coefficients in vehicle design \cite{arechiga2023drag, wang2024diffusebot}.

\subsection{Safety-Aware Machines}

Safety challenges are inherent in autonomous machines, where these robots must correct potential failures and explain their decisions to earn human trust. Ideally, a truly robust system would integrate multiple sensory inputs--visual, auditory, and tactile--so it can effectively detect failures, as certain cues are best identified by specific sensors. For example, auditory cues may detect a door slamming that visual sensors miss if they are obstructed. Like human drivers prioritizing important objects, robots must also extract key information from dense data while ignoring irrelevant details. It is also important that robotic systems are immune to the risks of out-of-distribution (OOD) input that may not be captured during simulation, as these can trick the system. For example, semantic anomalies (e.g., billboard stop signs) have been known to cause failures like autopilot disengagement or phantom braking \cite{elhafsi2023semantic}. To enhance safety, robots can incorporate various feedback mechanisms depending on the task. Human feedback through natural language corrections allows for real-time adjustments during long-horizon tasks where small errors may accumulate, while contact feedback is crucial in manipulation tasks to differentiate between desired contact (e.g., grasping) and harmful contact (e.g., collisions). 


Diffusion models have been used to generate OOD scenarios, enabling robots to train on rare and nuanced failure modes like compounding errors in long-horizon tasks \cite{zhang2024diffusion, sarva2023adv3d}. Risk-aware planning has improved with contact-awareness and future trajectory generation, allowing robots to navigate high-risk environments more reliably \cite{xie2023language, nishimura2023rap, danesh2023leader, cosner2024generative}. Multimodal models, which fuse data from visual, auditory, and tactile inputs alongside control signals, enhance safety by offering comprehensive explanations for robot actions and failures \cite{yuan2024rag, liu2023reflect, shao2023safety, renz2022plant}. Generative models also enable real-time human feedback, allowing robots to refine their actions and even request assistance when necessary \cite{shi2024yell, ren2023robots}. Finally, LLMs have strengthened anomaly detection, allowing robots to recognize and reason about semantic anomalies \cite{sinha2024real, elhafsi2023semantic}.

\subsection{Generalist Agents}

The pursuit of generalist agents marks a new frontier in robotics, where machines must seamlessly adapt to diverse environments, tasks, and hardware platforms. This adaptability addresses the limitations of current systems, which struggle to generalize across conditions \cite{zhu2023learning}. Robots are now building vast skill libraries, with LLM-guided bootstrapping enabling agents to expand their repertoires over time with minimal supervision \cite{zhang2023bootstrap}. Another key example is the Robotics Transformer (RT-1), a 35 million-parameter model trained on 700+ tasks and 130k demonstrations collected by 13 robots over 17 months, capable of generalizing to new tasks and scenes \cite{brohan2022rt}. Building on this, RT-2 pushes generalization further by combing web knowledge and robotic data in a  55 billion-parameter model \cite{brohan2023rt}. Octo further exemplifies versatility, trained on 800k trajectories across 9 platforms, and allows users to fine-tune within hours to new sensory inputs or action spaces \cite{team2024octo}. At an even larger scale, PaLM-E, a 562 billion-parameter model, integrates sensor modalities with language models, unlocking new possibilities for real-time interaction with the physical world \cite{driess2023palm}. Finally, RT-X, trained on data from 22 robots and 160,000+ tasks, shows improvements in generalization and highlights the impact of large-scale data collection \cite{padalkar2023open}. Together, these advancements signal the dawn of adaptable, general-purpose agents capable of transitioning seamlessly across tasks, environments, and platforms.

\section{Safety Challenges and Opportunities}

In virtual domains, chatbots and image generators are among the most common applications of generative AI, with ChatGPT alone serving over 180.5 million users \cite{deng2023early}. Much of the safety focus has understandably centered on these use cases. Common challenges include generating inappropriate content, harmful outputs, and amplifying social biases \cite{birhane2021multimodal, cho2023dall}. Additionally, interpretability issues arise, where models lack transparency in their outputs \cite{fui2023generative}. LLMs can also respond to harmful requests \cite{qi2023fine}, generate deep-fakes that compromise authenticity \cite{romero2024generative}, and inadvertently violate copyright laws \cite{chua2024ai}. Privacy risks arise from models that leak sensitive information \cite{weidinger2022taxonomy}, while hallucinations may lead to false or misleading information, particularly in fields such as law and medicine \cite{weidinger2022taxonomy}. Additional concerns include sycophancy, where models reinforce user misconceptions, and miscalibration, where they exhibit undue confidence in incorrect or outdated information \cite{liu2023trustworthy}.

In autonomous machines, many of these challenges persist and are often amplified due to the physical nature of tasks, while other challenges are unique to this domain. The following sections explore the key safety challenges of applying generative AI to autonomous machines, categorized into Model and System challenges.

\subsection{Model}

\subsubsection{Hallucinations}

\leavevmode

\textbf{Challenge}: Hallucinations are a major challenge with LLMs currently deployed in the virtual domain, where models confidently generate outputs that seem plausible but are incorrect and untethered from reality. Such false outputs pose amplified safety risks compared to virtual applications. For instance, an LLM-based planner \cite{ren2023robots} tasked with cooking could hallucinate that the stove is off when it is actually on, leading the robot to incorrectly handle a hot surface, potentially causing burns or even a fire hazard. These hallucinations can result in dangerous outcomes, highlighting the need for more robust generative models in safety-critical systems.

\textbf{Opportunities}: Developing systems that ask for help instead of hallucinating can greatly improve safety. In the earlier stove scenario, this would require the language model to recognize its uncertainty \cite{tanneru2024quantifying} and seek clarification (e.g., asking if the stove is on or off) \cite{ren2023robots}. Chain-of-thought (CoT) prompting, which encourages step-by-step task breakdowns, is another promising approach for improving reasoning \cite{wei2022chain}. However, it is important that CoT explanations faithfully reflect the model's true reasoning. Studies have shown that while LLMs can generate human-appealing explanations, these often misrepresent the true reason for a model’s prediction \cite{tanneru2024difficulty, lanham2023measuring, turpin2024language}, leading to further safety concerns. Other promising techniques are also being developed that can help reduce hallucinations in LLM-based systems, such as real-time verification and rectification which validates outputs on the fly making sure that the output is grounding in reality \cite{tonmoy2024comprehensive}. 

\begin{figure}[t]
    \centering
    \includegraphics[width=\linewidth]{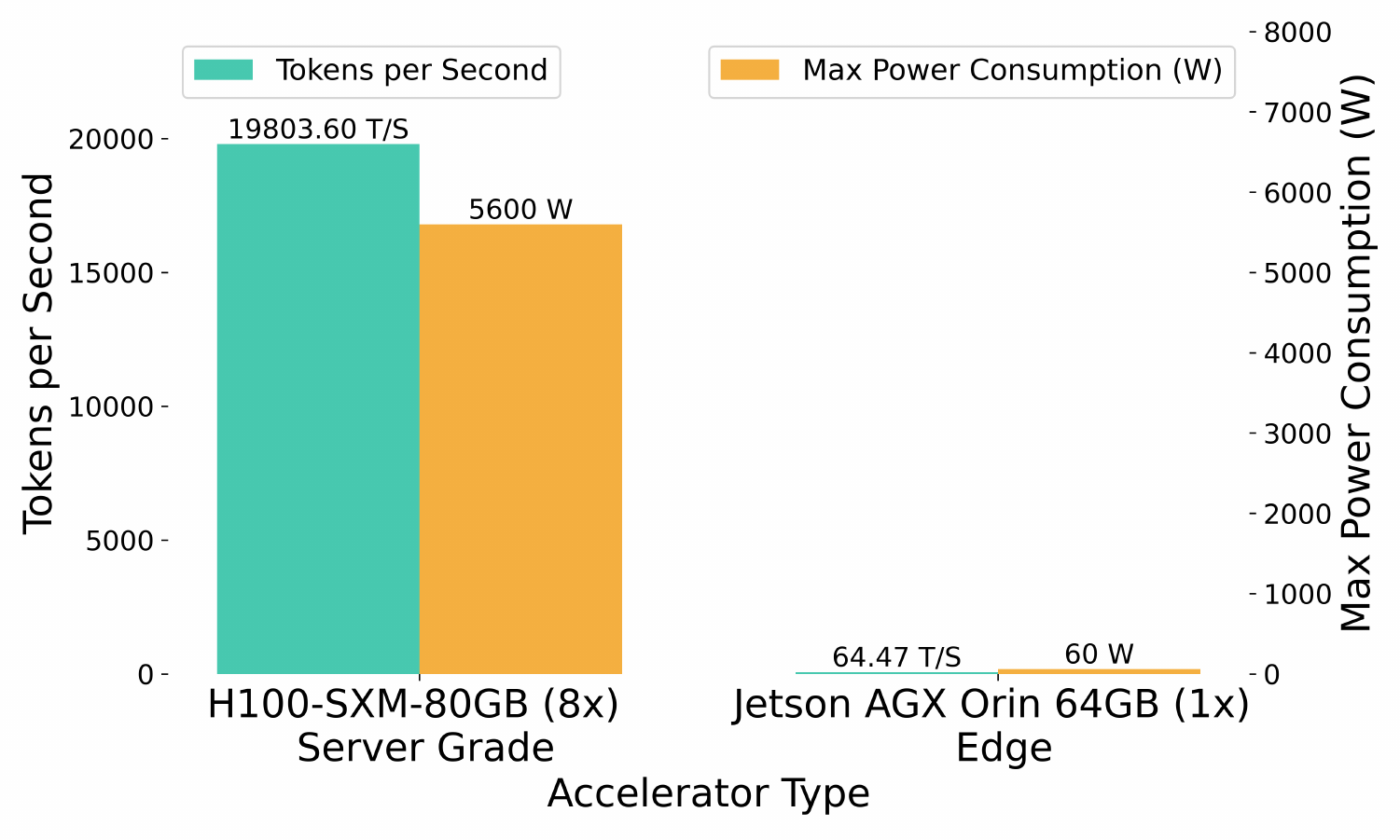}  
    \caption{Server-grade setup (8 NVIDIA H100 GPUs with Intel(R) Xeon(R) Platinum 8468) versus a common edge setup for robotics (NVIDIA Jetson AGX Orin) from the MLPerf Inference benchmarks on the GPT-J 6B model for the LLM summarization task. It shows the discrepancy in tokens/second and power consumption \cite{mlcommons_benchmark, nvidia_jetson_orin, hyperstack_h100_benchmarks}.}
    \label{fig:h100_vs_orin}
    \vspace{-1em}
\end{figure}

\subsubsection{Catastrophic Forgetting}
\label{sec:cat_forg}
\leavevmode

\textbf{Challenge}: In robotic systems that fine-tune LLMs for specific tasks, catastrophic forgetting can occur, where improvements on the fine-tuned task come at the cost of forgetting broader knowledge learned during pretraining. Since fine-tuning datasets are smaller and less diverse, models risk losing critical capabilities from pretraining, and it may not always be clear what the model has forgotten, posing a safety risk as the system might unknowingly be incapable of reasoning about certain situations \cite{zhai2023investigating}. Alternatively, another approach to teach LLMs new tasks is through in-context learning, which occurs during inference without updating model weights, allowing for fast adaptation to language instructions. However, this method is constrained by limited context windows, such as LLaMA2’s 4096 token limit \cite{touvron2023llama}, which restricts in-context learning to short-term interactions. In robotics, where human feedback is beginning to play a critical role in guiding behavior, this limitation poses a safety risk. If human instructions accumulate over long multi-step tasks and fall outside the context horizon, critical commands can be forgotten, leading to unsafe actions \cite{liang2024learning}.

\textbf{Opportunities}: Current work is focused on reducing the amount of pretrained capabilities lost during fine-tuning, such as Conjugate Prompting, which artificially makes the task appear farther from the fine-tuning distribution \cite{kotha2023understanding}. Additionally, efforts are underway to better understand the capabilities of in-context learning, as this emergent paradigm remains not fully understood \cite{garg2022can, wies2024learnability}. Techniques like Rotary Positional Embeddings (RoPE) \cite{su2024roformer} and LongROPE \cite{ding2024longrope} are being developed to extend context windows and help models retain longer sequences. However, these methods increase system memory usage, which could pose challenges for edge devices, as discussed in Section \ref{sec:resource-requirements}.

\subsubsection{Lack of Formal Guarantees}

\leavevmode

\textbf{Challenge}: Generative models lack formal guarantees, meaning they provide no mathematically-proven assurances about stability, safety, or performance. In contrast, traditional robotics algorithms, such as PID controllers, LQR, and MPC, offer formal guarantees of stability (e.g., the system will converge to a desired state or maintain a trajectory) and performance (e.g., minimizing error over time). Similarly, motion planning methods like RRT and A* provide guarantees of completeness (if a solution exists, the algorithm will find it) or optimality (the solution is the best possible under the given constraints). These approaches ensure predictable behavior. Generative models, however, rely on training data and probabilistic reasoning, which can lead to unpredictable outcomes, especially in unfamiliar or OOD scenarios. This probabilistic nature makes it challenging to provide deterministic safety or stability when replacing traditional algorithms with generative models.




\textbf{Opportunities}: There is potential in hybrid approaches that integrate control theory methods like Control Barrier Functions (CBFs) and Control Lyapunov Functions (CLFs) to enforce safety and stability in generative AI systems. CBFs construct safe control constraints to prevent unsafe states, while CLFs ensure stable goal convergence. Recent work applies these methods in areas like constraint-aware diffusion models \cite{li2024constraint}, CBF-inspired interventions for safe LLM outputs \cite{miyaoka2024cbf}, and diffusion models leveraging both CBFs and CLFs to enforce safety and stability properties \cite{mizuta2024cobl}.

\begin{figure}[t]
    \centering
    \includegraphics[width=\linewidth]{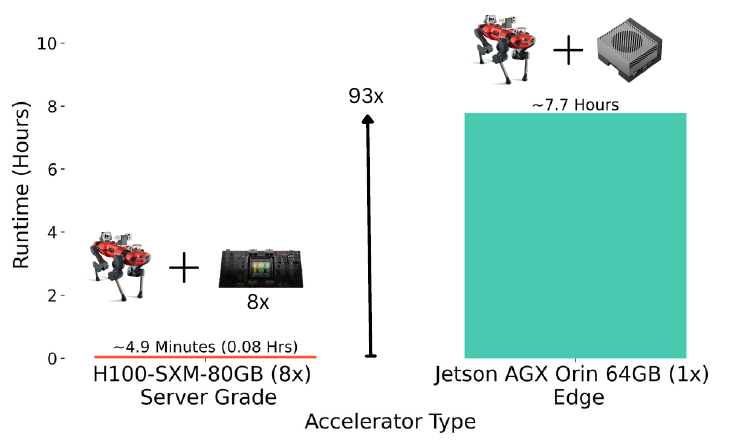}  
    \caption{Max runtime comparison of an 8 NVIDIA H100 GPU setup versus a NVIDIA Jetson AGX Orin on a 932.4 Wh battery pack of an ANYmal quadruped \cite{anymal_specs}, conservatively assuming actuators only consume 50\% of the power \cite{wu2023review} and 100\% FLOP utilization is achieved. This highlights the infeasibility of achieving server-grade throughput on an edge platform due to battery limitations.}
    \label{fig:runtime}
    \vspace{-1em}
\end{figure}

\subsubsection{Lack of Real-World Grounding}

\leavevmode

\textbf{Challenge}: A significant weakness of language models is their lack of real-world grounding, meaning they were not trained to directly perceive or interact with the physical world, making it difficult to apply them for decision-making in specific robotic embodiments \cite{rana2023sayplan}. For example, asking a language model to describe how to clean a spill might generate a reasonable narrative, but it may not be applicable to a particular robot morphology \cite{ahn2022can}. This lack of physical experience is a key reason why techniques developed in areas like visual question answering (VQA) and view-based navigation have not transitioned smoothly into embodied agents \cite{biggie2023tell}. Bridging this gap requires extensive robotics data, which is costly and time-consuming. For instance, while ChatGPT-3 was trained on 300 billion tokens of web text, collecting the data for RT-1 required 13 robots and 17 months to gather just 130k episodes—a significant amount, but far smaller in scale than the data used for large language models \cite{brohan2022rt}. Without sufficient data, robots are more prone to OOD scenarios, increasing the risk of unsafe behavior.

\textbf{Opportunities}: The field is advancing toward building generalist models by pooling diverse robotics data across platforms. RT-X is at the forefront, aggregating data from 34 labs across 22 embodiments, including single-arm robots, bi-manual systems, and quadrupeds \cite{padalkar2023open}. While this marks a significant step forward, much more work is needed to expand data collection across an even wider range of robotic platforms and environments. These efforts lay the groundwork for the creation of Large Robot Models (LRMs)—the next major leap in robotics.

\begin{figure}[t]
    \centering
    \includegraphics[width=\linewidth]{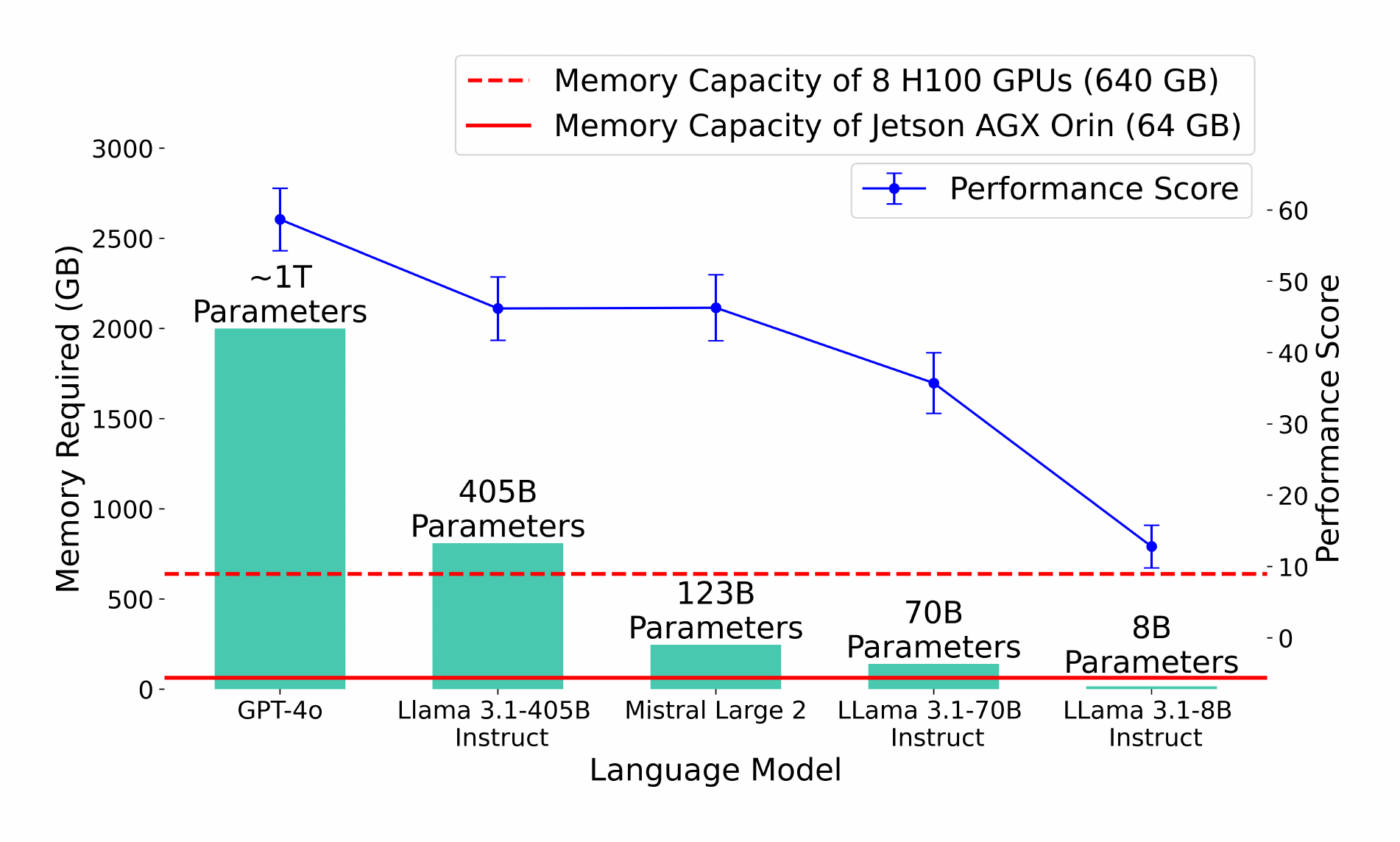}  
    \caption{Memory required to load LLMs (assuming FP16 parameters) and benchmarked step-by-step reasoning performance \cite{scale_tool_use}. The Jetson AGX Orin is limited to small models (e.g. LLama 3.1-8B), due to memory constraints-- thus impacting reasoning capabilities.}
    \label{fig:memory}
    \vspace{-1em}
\end{figure}

\subsection{System}

\subsubsection{Real-Time Processing}

\leavevmode

\textbf{Challenge}:
Diffusion models are gaining popularity as a powerful tool for motion planning \cite{stamatopoulou2024dippest}, but their inherent latency poses a significant challenge. The high latency is due to the repeated inference steps required for denoising. For example, 2D path planning using diffusion operates at 2.5 Hz \cite{liu2024dipper}, while autonomous vehicles typically require path planning at 20 Hz \cite{saba2024fast}, or even faster in dense, safety-critical environments. Such decision frequencies fall short of real-world requirements, raising safety concerns due to delayed response times in critical situations.

The growing scale of LLMs also hinders real-time, reactive reasoning in mobile robots, particularly when they are deployed on edge devices. For instance, common edge devices for robots, like the NVIDIA Jetson AGX Orin, offer drastically lower compute power compared to server-grade GPUs such as the NVIDIA H100, which are used for virtual applications like chatbots. As shown in Figure \ref{fig:h100_vs_orin}, 8 NVIDIA H100 GPUs can process $307\times$ more tokens per second than the NVIDIA Jetson AGX Orin. While deploying multiple GPUs is common in server environments, edge devices typically lack the ability to parallelize across multiple accelerators. As a result, systems like the Orin must operate individually, highlighting the vast disparity in both compute power and scalability between edge devices and server-grade systems. The Orin’s performance is significantly slower, underscoring the challenge of even running small language models on edge devices, as these values come from benchmarking a relatively small language model—GPT-J with only 6 billion parameters. Larger models are often too resource-intensive to deploy on edge devices, leading to reduced reasoning capabilities. Large generalist models for motion planning, like RT-2, can only run at 1-3 Hz \cite{brohan2022rt}. This low frequency is exacerbated by the need to run on edge devices, since alternative deployment options such as running inference on the cloud typically introduces communication latency of more than 100 ms- far exceeding the 10-100 ms required for many robotic applications \cite{qu2024mobile}.




\textbf{Opportunities}: Efforts like distilling student networks (e.g., Consistency Policy \cite{prasad2024consistency}) and simplifying models during denoising \cite{dong2024diffuserlite} have shown promise in speeding up diffusion models. Language models are also being optimized through better parameter utilization, KV cache management, and parallel decoding \cite{huang2024new}. However, many systems have poor Machine FLOPs Utilization (MFU), typically 50\% or less. Significant opportunities lie in maximizing MFU to fully exploit computational resources, alongside domain-specific optimizations that leverage the unique constraints of robotics, beyond what’s achievable in virtual environments.

\subsubsection{Resource Requirements}
\label{sec:resource-requirements}

\leavevmode

\textbf{Challenge}: Deploying generative models demands significant resources, especially in memory and power consumption. Server-grade setups like 8 NVIDIA H100 GPUs consume up to 5600 W, whereas edge devices such as the NVIDIA Jetson AGX Orin use just 60 W (Figure \ref{fig:h100_vs_orin}). Edge devices are more power-efficient but sacrifice computational strength. Achieving server-grade throughput on the edge would require much larger batteries than typically available. For example, as shown in Figure \ref{fig:runtime}, the Jetson AGX Orin—commonly used in robots like the ANYmal quadruped \cite{jenelten2022tamols}—can operate for several hours, while the same battery powering 8 NVIDIA H100 GPUs would last only a few minutes. This highlights the infeasibility of achieving server-level throughput on the edge due to power constraints, leading to high-latency inference and impacting safety.

From a memory standpoint, LLMs require significant space just to store their weights (Figure \ref{fig:memory}). The Jetson AGX Orin, due to limited memory, can only run smaller LLMs, leading to performance degradation, as Figure \ref{fig:memory} illustrates. Generally, smaller models result in lower performance, meaning LLMs deployed on edge devices for robotic applications will have inferior reasoning capabilities compared to those running on servers for virtual agents. This reduction in reasoning ability compromises safety in critical tasks.

Memory limitations also make addressing catastrophic forgetting (Section \ref{sec:cat_forg}) even more difficult. Transformer models, in particular, face quadratic memory scaling with context window size. As larger context windows are vital for improved reasoning and decision-making, memory restrictions on edge devices hinder deploying robust models capable of long-term retention and adaptation.


\textbf{Opportunities}: Quantization and pruning offer ways to reduce power and memory usage, enabling LLMs to run on more resource-constrained hardware. Quantization shrinks model size by lowering precision, while pruning cuts unnecessary components, reducing computational demands. Distilling LLM knowledge further eases resource constraints by minimizing reliance on full models \cite{ginting2024seek}.


\section{Toward a Safety Scorecard for Risk Assessment}

As generative models become increasingly integrated into physical agents, we face an urgent need to quantify and clearly communicate associated risks to all stakeholders. Our exploration has shown how these models can enhance various levels of the autonomous stack, from context-aware planning to human-robot communication. However, this diversity in application brings with it a spectrum of safety challenges, with risk levels varying significantly based on the specific integration within the robotic system.


We propose a critical next step: the development and adoption of a standardized safety scorecard for autonomous systems that incorporate generative AI. Drawing inspiration from model cards \cite{mitchell2019model}, which document the intended use cases of machine learning models, our proposed framework goes further. It aims to provide a comprehensive and standardized method for rating and communicating safety risks, tailored specifically to where and how generative models are integrated into the autonomous stack.


\textit{Case Study}: We consider an autonomous system with generative models integrated at four levels of the computing stack: an LLM for visual anomaly detection \cite{elhafsi2023semantic}, an LLM for improving explainability by describing and justifying vehicle actions \cite{yuan2024rag}, a 2D diffusion-based path planner \cite{stamatopoulou2024dippest}, and an LLM for personalizing the autonomous driving experience using human instructions \cite{cui2024receive}. 

Using the MLCommons risk rating scale \cite{mlcommons_ai_safety_benchmark_2024}, we show how this scorecard could work. For example, Figure \ref{fig:scorecard} shows the explainability system ranked Low Risk, even with occasional hallucinations, as it does not directly affect vehicle control. In contrast, the anomaly detection system is classified as High Risk due to its low latency and the potential for missed anomalies leading to crashes.




\definecolor{highrisk}{HTML}{F8C78C}
\definecolor{lowrisk}{HTML}{D0ECD6}
\definecolor{moderatehighrisk}{HTML}{FBE1C1}

\begin{figure}[t]
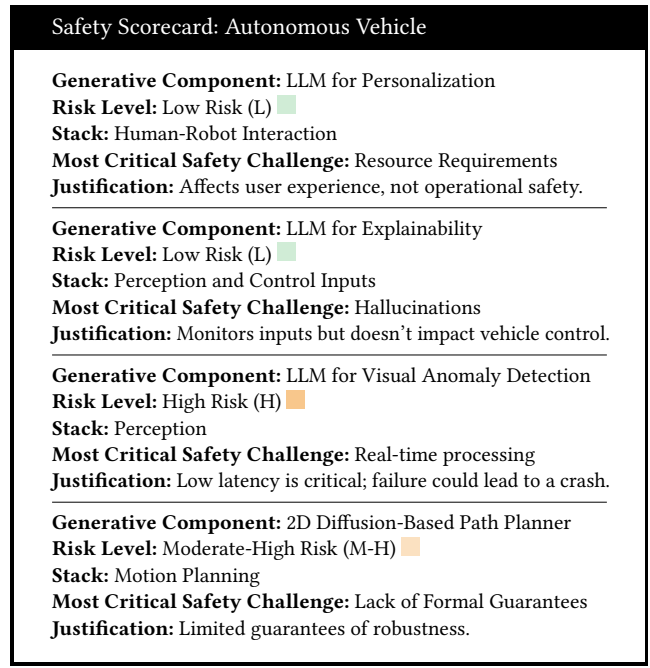

    \centering

    \begin{tcolorbox}[colframe=black,colback=white!10,sharp corners, width=\linewidth, title=Safety Scorecard: Autonomous Vehicle]

    \small
    \textbf{Generative Component:} LLM for Personalization \\
    \textbf{Risk Level:} Low Risk (L)~\textcolor{lowrisk}{\rule{2ex}{2ex}}\\
    \textbf{Stack:} Human-Robot Interaction \\
    \textbf{Most Critical Safety Challenge:} Resource Requirements \\
    \textbf{Justification:} Affects user experience, not operational safety.
    
    \vspace{-0.15cm}
    \rule{\linewidth}{0.3pt}
    
    \textbf{Generative Component:} LLM for Explainability \\
    \textbf{Risk Level:} Low Risk (L)~\textcolor{lowrisk}{\rule{2ex}{2ex}} \\
    \textbf{Stack:} Perception and Control Inputs \\
    \textbf{Most Critical Safety Challenge:} Hallucinations \\
    \textbf{Justification:} Monitors inputs but doesn’t impact vehicle control. 

    \vspace{-0.15cm}
    \rule{\linewidth}{0.3pt}

    \textbf{Generative Component:} LLM for Visual Anomaly Detection \\
    \textbf{Risk Level:} High Risk (H)~\textcolor{highrisk}{\rule{2ex}{2ex}}  \\
    \textbf{Stack:} Perception \\
    \textbf{Most Critical Safety Challenge:} Real-time processing \\
    \textbf{Justification:}  Low latency is critical; failure could lead to a crash.
    
    \vspace{-0.15cm}
    \rule{\linewidth}{0.3pt}
    
    \textbf{Generative Component:} 2D Diffusion-Based Path Planner \\
    \textbf{Risk Level:} Moderate-High Risk (M-H)~\textcolor{moderatehighrisk}{\rule{2ex}{2ex}}\\
    \textbf{Stack:} Motion Planning \\
    \textbf{Most Critical Safety Challenge:} Lack of Formal Guarantees \\
    \textbf{Justification:} 
    Limited guarantees of robustness.
    
    \end{tcolorbox}
    \vspace{-1em}
    \caption{Example of a ``safety scorecard'' for generative models across four levels of the computing stack in an autonomous system.}
    \label{fig:scorecard}
    \vspace{-2em}
\end{figure}

\section{Conclusion}
Generative AI expands existing capabilities and unlocks new potential in autonomous machines, but its integration into physical systems raises critical safety concerns. To this end, we have examined the various applications of generative AI in robotics, highlighting both its transformative power and the unique safety challenges it presents. We conclude by issuing an urgent call to action for the development of AI safety scorecards specifically designed for generative AI in autonomous machines. These scorecards will serve as a standardized framework for assessing and communicating the safety risks associated with integrating generative models at different levels of the autonomous stack. 
\begin{acks}
This work is supported by the National Science Foundation (NSF) GRFP under Grant No. DGE-2140743. Any opinions, findings, conclusions, or recommendations expressed in this material are those of the authors and do not necessarily reflect the views of the NSF.
\end{acks}

\bibliographystyle{ACM-Reference-Format}
\bibliography{references}

\end{document}